\documentclass[10pt,twocolumn,letterpaper]{article}

\usepackage{wacv}
\usepackage{times}
\usepackage{epsfig}
\usepackage{graphicx}
\usepackage{amsmath}
\usepackage{amssymb}
\usepackage{booktabs}
\usepackage{tabularx}
\usepackage{multirow,bigdelim}
\usepackage{enumitem}
\usepackage[font=footnotesize,labelfont=bf]{caption}

\newcommand{\figref}[1]{Fig.~\ref{fig:#1}}
\newcommand{\tblref}[1]{Table~\ref{tbl:#1}}
\newcommand{\secref}[1]{Section~\ref{sec:#1}}

\usepackage{subcaption}
\newcommand{\figgallery}{
\begin{figure}[t]
   \centering    
   \includegraphics[width=1\linewidth]{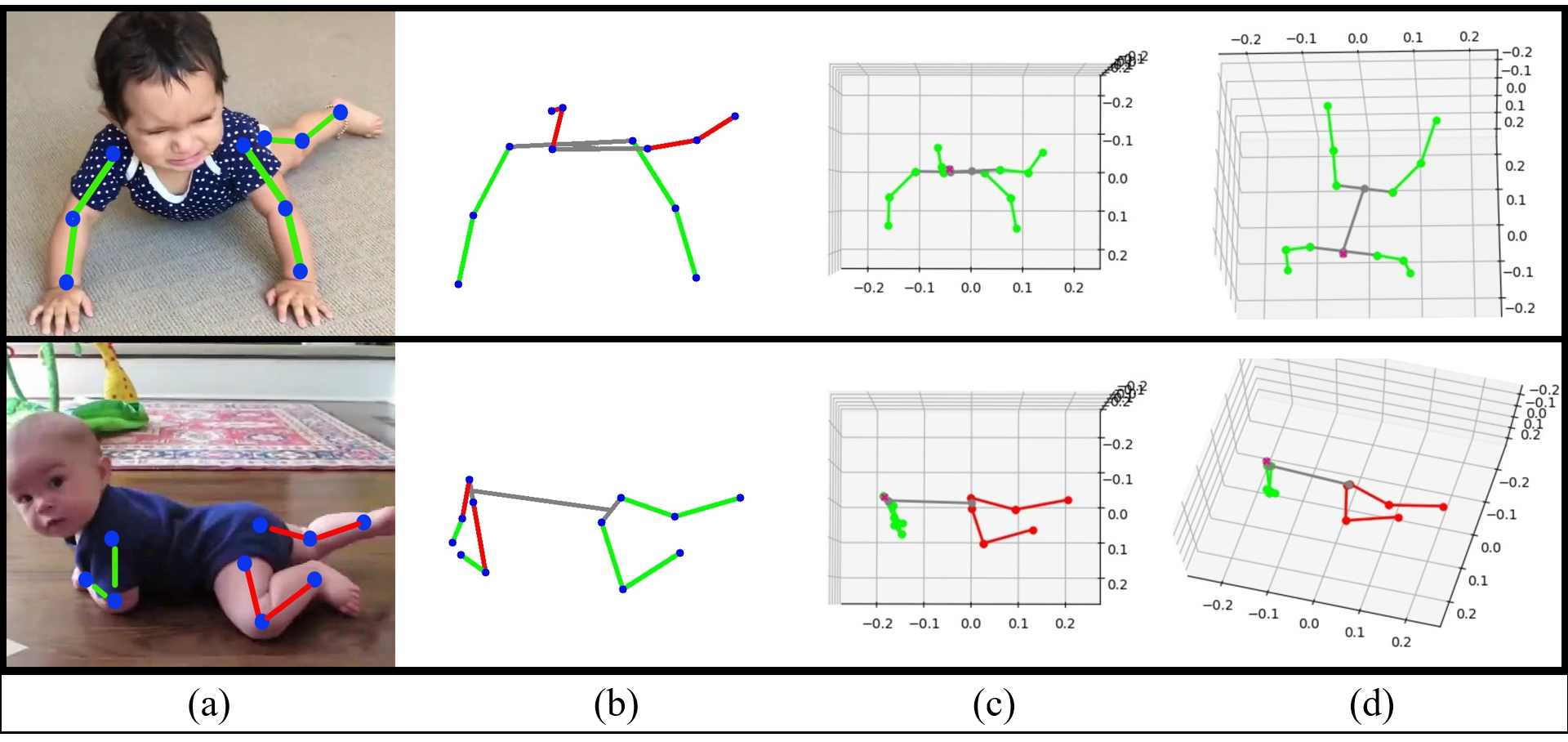}
    \vspace{-.2in}
    \caption{Examples of discrepancy between 2D pose-based and 3D pose-based symmetry measurements. Four limb pairs are annotated in different colors based on the corresponding symmetry label. If the limb pair is symmetric, both side of the limb parts are marked in green, otherwise, they are in red. Except for these four limb pairs, other parts of the body skeleton are uniformly plotted in gray. (a) Bayesian aggregated symmetry result from human ratings as weak ground truth on the raw image (the occluded limb parts are not shown). (b) The 2D posed-based measurement results on 2D predicted skeleton. (c) and (d) the 3D posed-based measurement results on 3D predicted skeleton under two different viewing angles.} 
    \label{fig:gallery}
    \vspace{-.2in}
\end{figure}
}

\newcommand{\figsymmthd}{
\begin{figure}[t]
   \centering    \includegraphics[width=0.8\linewidth]{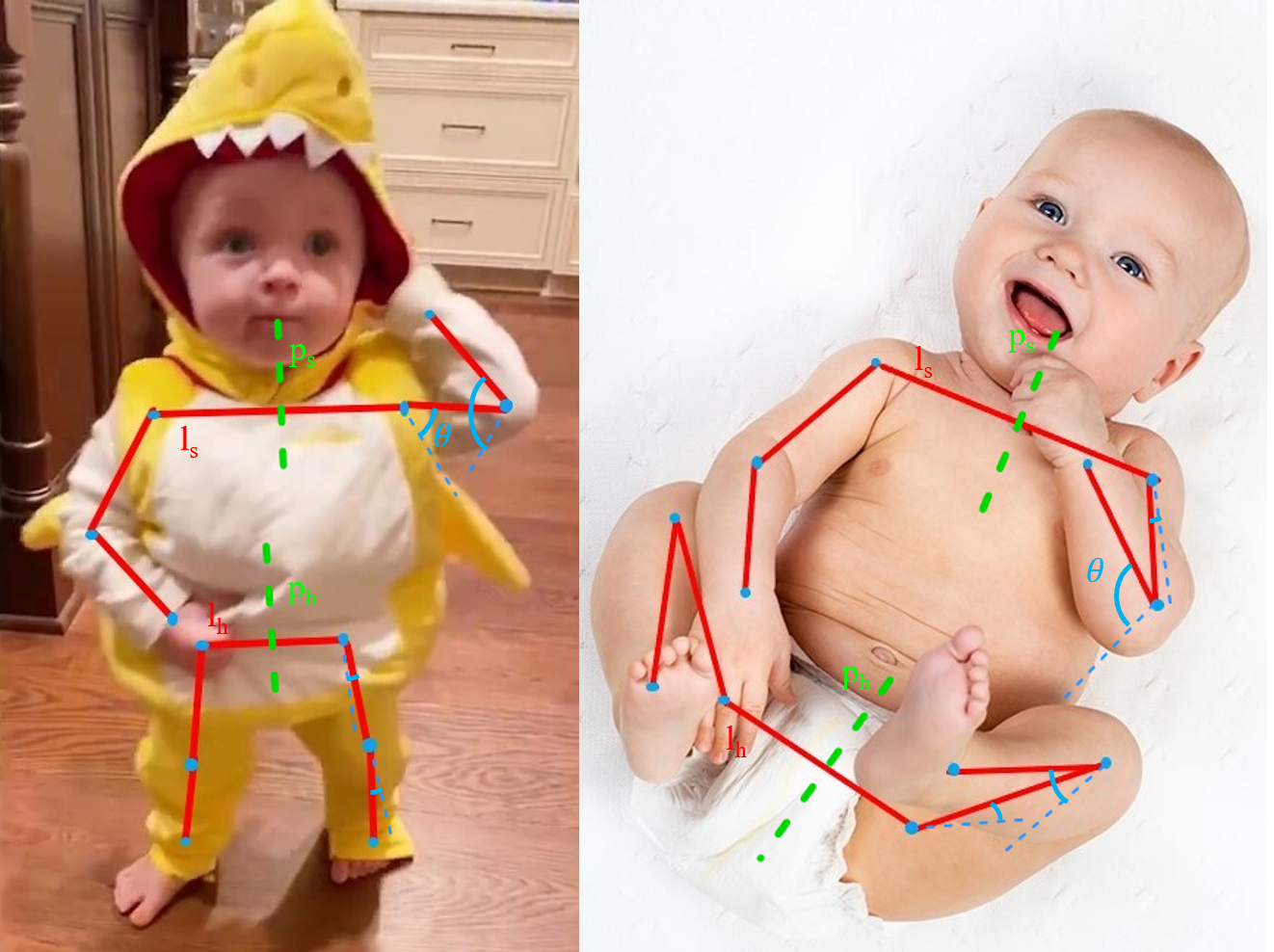}
    \vspace{-.11in}
    \caption{The illustration of our infant pose symmetry measurement. The right upper arm and lower arm in red are mirrored across the mid-perpendicular line $p_\text{s}$ (in 2D) or plane (in 3D) of the two shoulder joints in green, aligned with their left counterparts at the root joints, resulting in the phantom limbs in blue. The right upper and lower legs are likewise mirrored across the mid-perpendicular $p_\text{h}$ of the two hip joints and aligned with their left counterparts. All four resulting angles $\theta$ are measured, and the limb pair is considered pose symmetric if the its calculated angle is less than a predefined threshold.}
    \label{fig:symmthd}
    \vspace{-.2in}
\end{figure}
}

\newcommand{\figrateragreement}{
\begin{figure*}[t]
   \centering    
   \includegraphics[width=\linewidth]{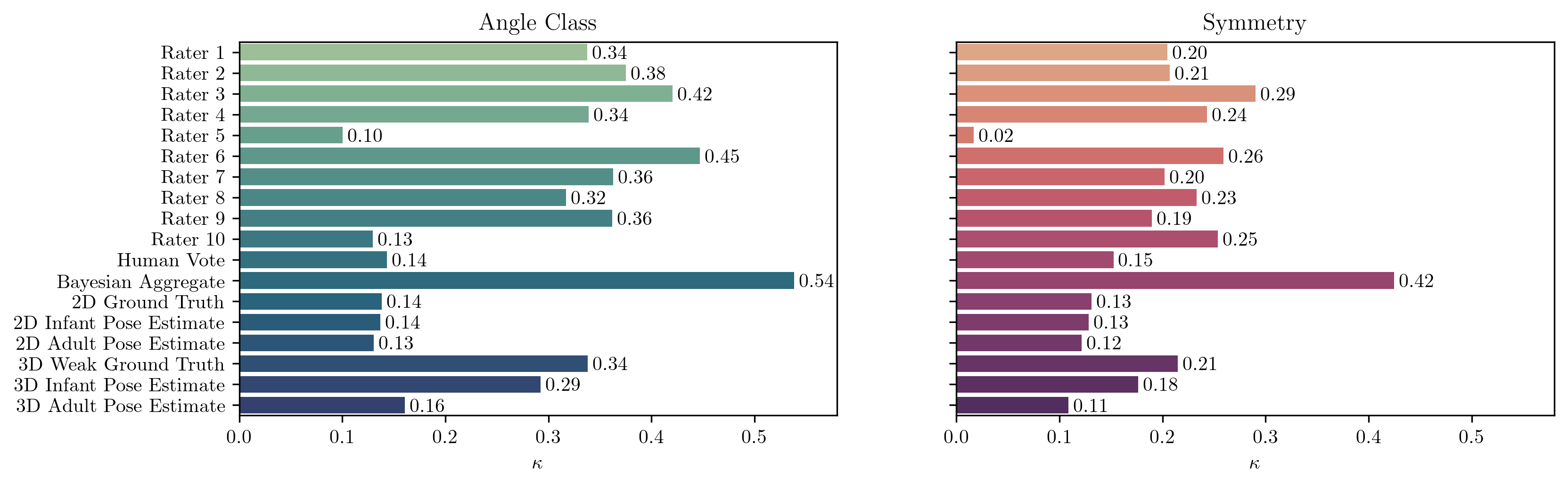}
    \vspace{-.3in}
    \caption{Average Cohen's $\kappa$ agreement of a given individual assessment with the 10 human rater assessments (with self-agreement excluded for the human raters). Among human raters, Raters 5 and 10 stand out as outliers for angle assessment, as does Rater 5 for symmetry assessments. The Bayesian aggregate assessment exhibits high average agreement, as expected, but interestingly the human-voted assessment does not. Among the pose-based assessments, those derived from 3D ground truth or 3D predicted poses by using HW-HuP model agree most strongly with human assessments, especially for the more objective assessment of angle level. Since the angle class is ordered, we employ the quadratically weighted Cohen's $\kappa$ for those assessments.} 
    \label{fig:rater-agreement}
    \vspace{-.2in}
\end{figure*}
}

\newcommand{\figpartcorrelations}{
\begin{figure}[t]
   \centering    
   \includegraphics[width=\columnwidth]{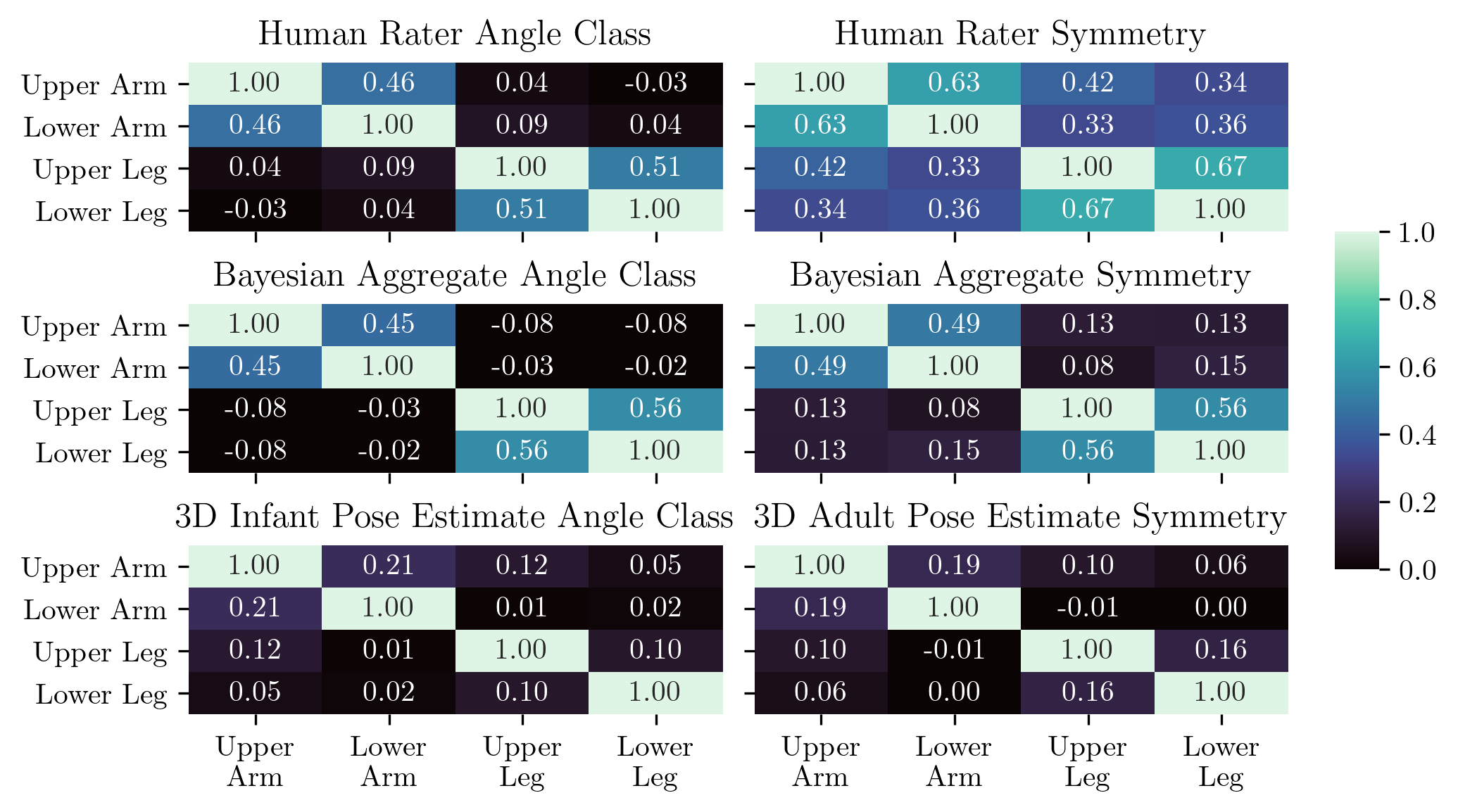}
    \vspace{-.25in}
    \caption{Cohen's $\kappa$ agreements between limbs of human, Bayesian aggregate, and 3D pose estimation, and 3D weak ground truth based assessments for angle and symmetry. While some level of inter-limb agreement may exist in the actual (inaccessible) ground truth data, the high agreements in human assessment of symmetry seem particularly excessive, and likely attributable to bias.} 
    \label{fig:part-correlations}
    \vspace{-.1in}
\end{figure}
}

\newcommand{\figinternalconsistency}{
\begin{figure}[t]
   \centering    
   \includegraphics[width=\columnwidth]{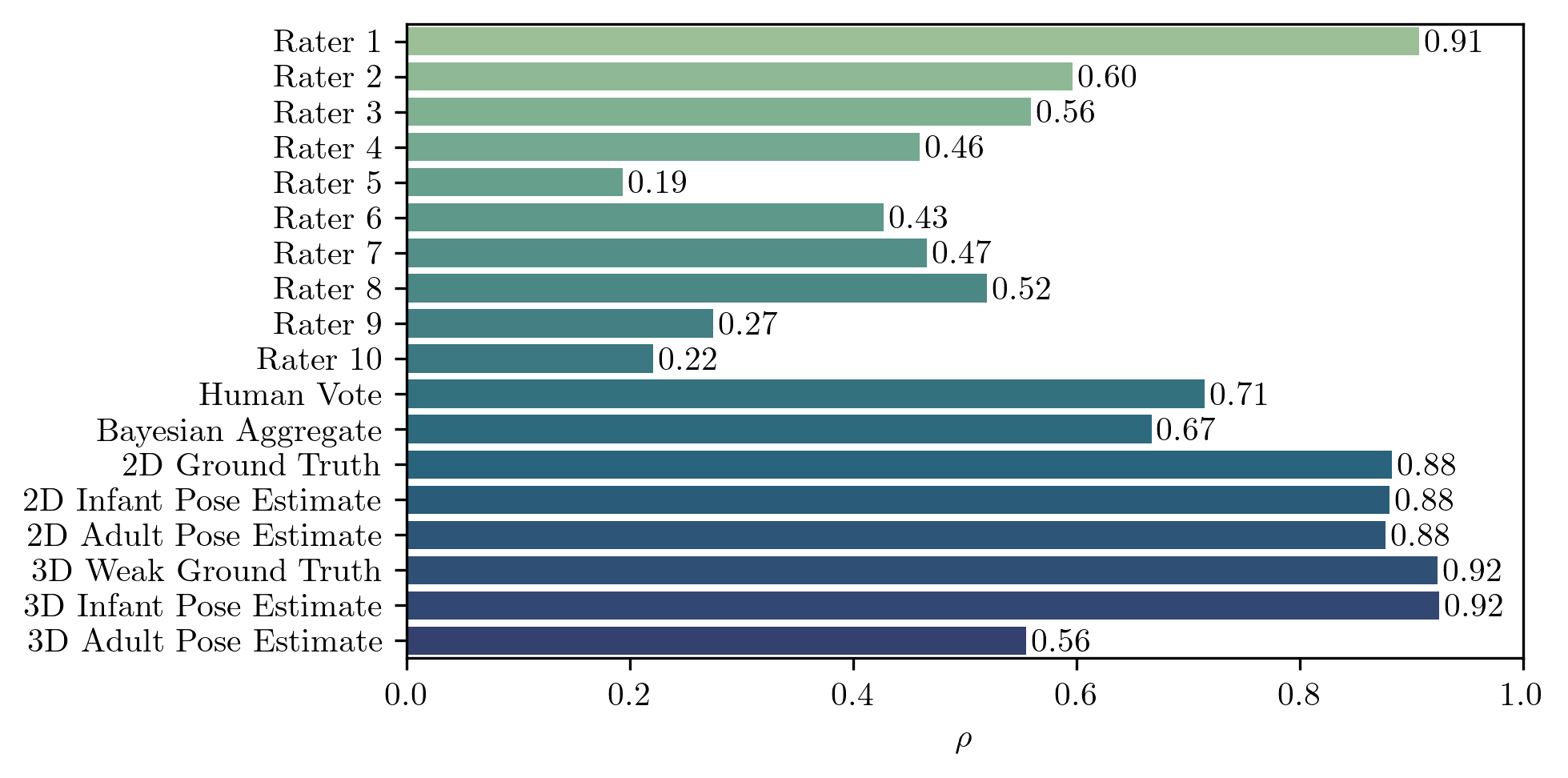}
    \vspace{-.3in}
    \caption{Spearman's $\rho$ ranked correlation between each assessor's angle and asymmetry assessments across all infants and limb pairs. Assessments with high scores can be interpreted as enjoying high ``internal consistency.'' Low scores can be caused either by low internal consistency or by angle threshold misalignment (as with the 3D adult pose-based model).} 
    \label{fig:internal-consistency}
    \vspace{-.2in}
\end{figure}
}

\newcommand{\figroccurve}{
\begin{figure}[t]
   \centering    
   \includegraphics[width=\linewidth]{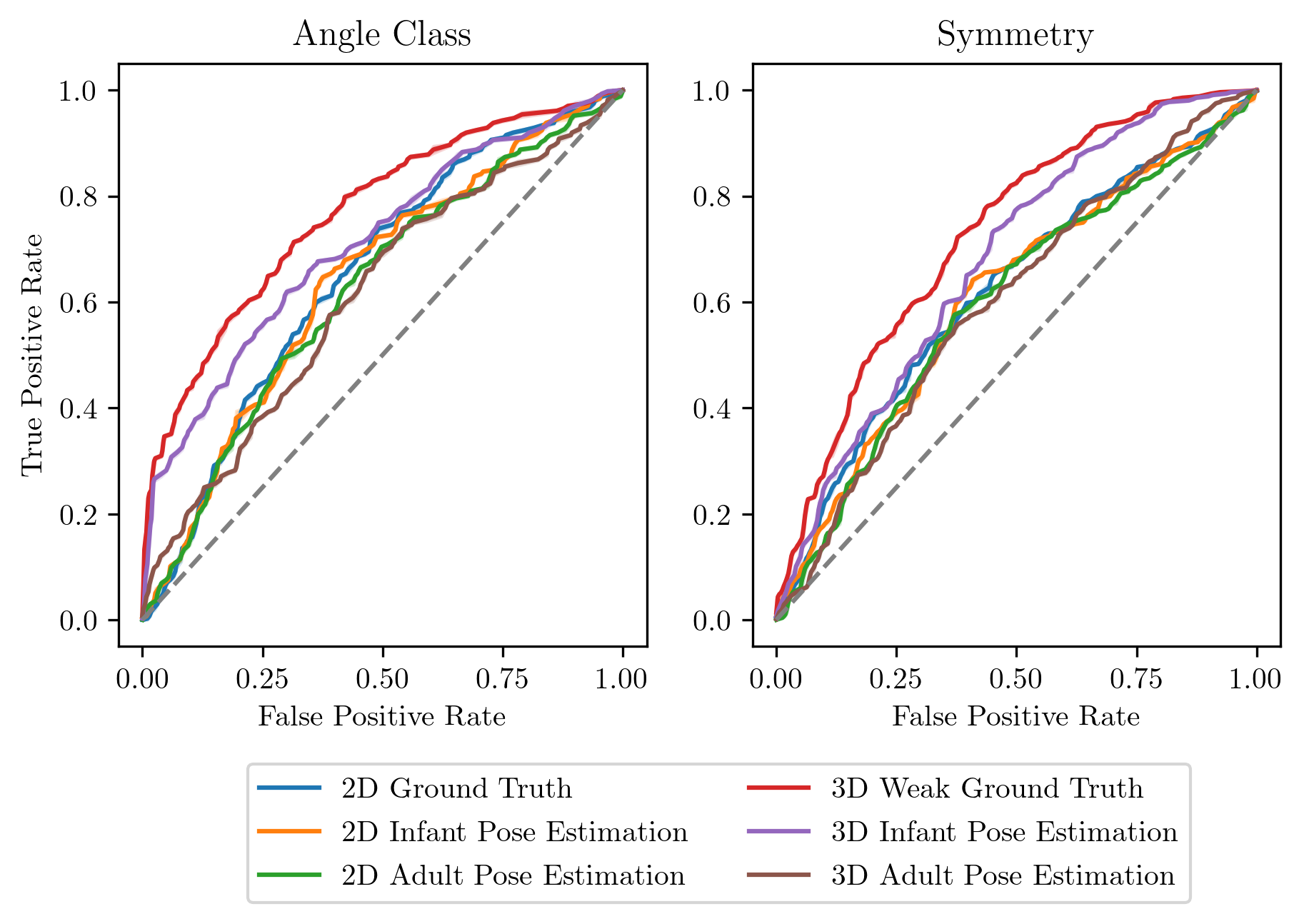}
    \vspace{-.2in}
    \caption{Receiver operating characteristic (ROC) curves indicating test set performance of logistic regression, of the Bayesian aggregate angle class and symmetry, from raw angles derived from the indicated pose estimation models. The regression based on our weak 3D ground truth yields best results, while the 3D infant pose estimation model performs better than the other estimation models or the 2D ground truth. The angle classes have been compressed into two, [${<}30^\circ$, ${\geq}30^\circ$], for simplicity. Corresponding areas under the curve (AUCs) can be found in \tblref{estimators}.} 
    \label{fig:roc-curve}
    \vspace{-.2in}
\end{figure}
}

\newcommand{\figoptimalangle}{
\begin{figure*}[t]
   \centering    
   \includegraphics[width=\linewidth]{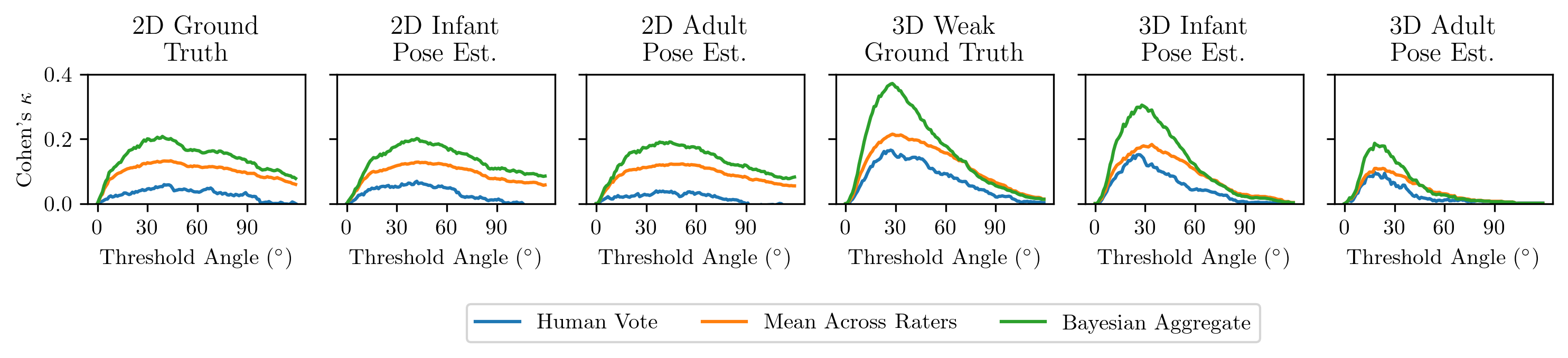}
    \vspace{-.20in}
    \caption{Cohen's $\kappa$ agreement between various pose estimation based assessments of symmetry and human assessments, as the threshold angle for defining the threshold for pose base assessment varies. Agreement with human assessment is either given as the mean of agreement with each of the 10 raters, or as agreement with the voted or Bayesian aggregate rater. These results show that the highest capacity for agreement is afforded by the Bayesian aggregate rater on the one hand, and the 3D infant pose estimation based model on the other.}
    \label{fig:optimal-angle}
    \vspace{-.1in}
\end{figure*}
}

\newcommand{\figtypical}{
\begin{figure}[t]
   \centering    
   \includegraphics[width=\linewidth]{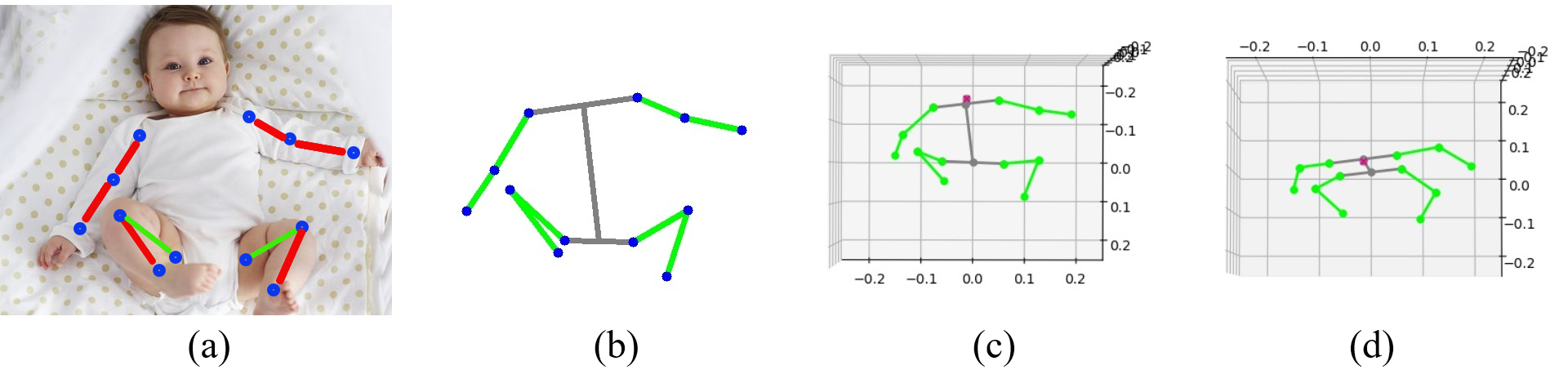}
    \vspace{-.2in}
    \caption{A special example demonstrating the effect of angle threshold on pose-based measurement results and constraints on the feasibility of 2D pose-based method. Layout and labels as in \figref{gallery}.
 } 
    \label{fig:typical}
\end{figure}
}

\newcommand{\figraterthresholds}{
\begin{figure*}[b]
   \centering    
   \includegraphics[width=\linewidth]{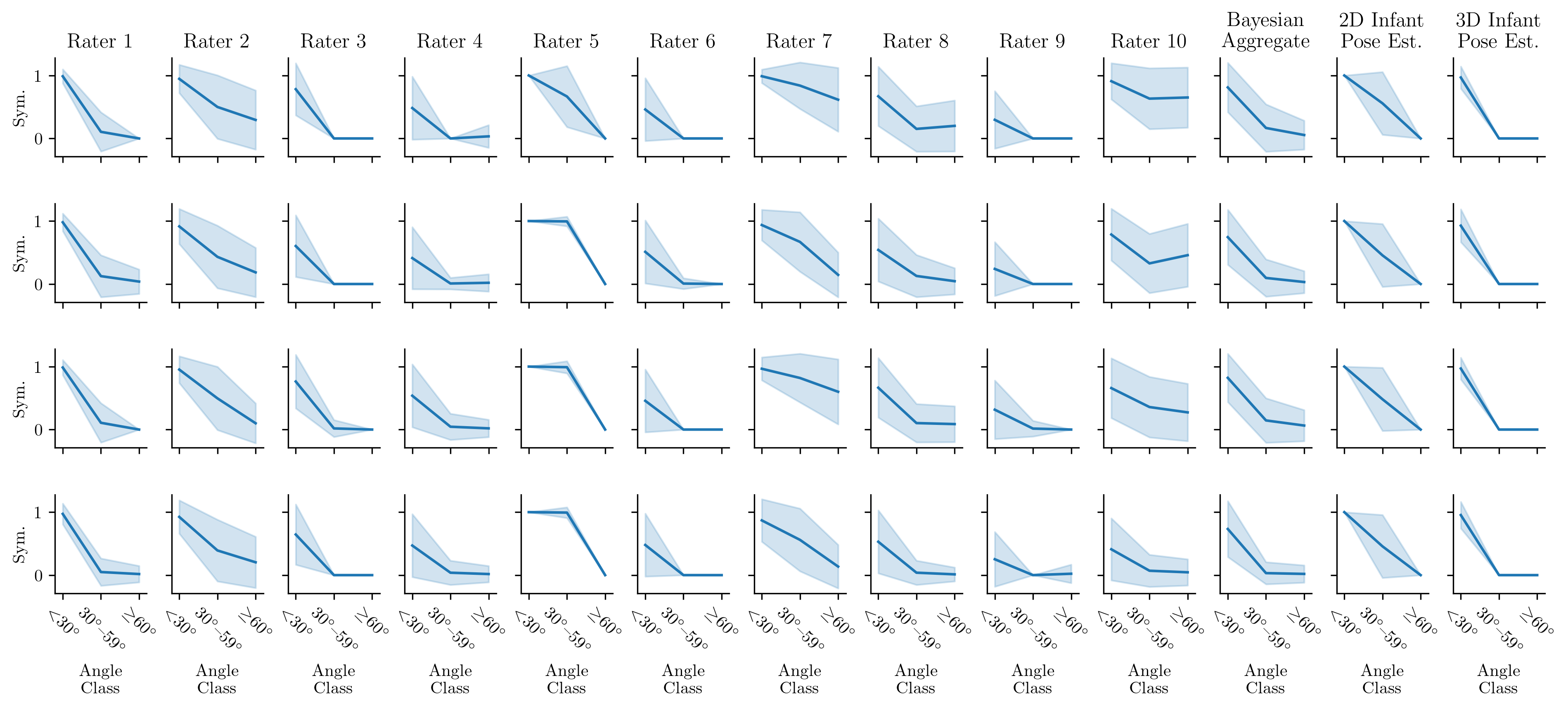}
    \vspace{-.11in}
    \caption{Mean assessments of symmetry level by assessed angle difference level, for all 10 raters plus the Bayesian aggregate rater and the predicted 2D- and 3D-pose-based models. Means are taken over all 700 SyRIP infant images per each of four pairs of limbs, with confidence intervals of one standard deviation at each angle level indicated. These statistics reveal wide variance in determination of symmetry versus angle difference across human raters, although most raters are fairly consistent across limbs. An upwardly sloped segment, as seen most prominently in Rater 8's upper arm assessments and or Rater 10's lower arm assessments, indicates an apparent inconsistency in aggregate assessments. Note that while small confidence intervals indicate consistent assessments, the converse does not necessarily hold.} 
    \label{fig:rater-thresholds}
\end{figure*}
}

\newcommand{\figanglehistograms}{
\begin{figure*}[b]
   \centering    
   \includegraphics[width=\linewidth]{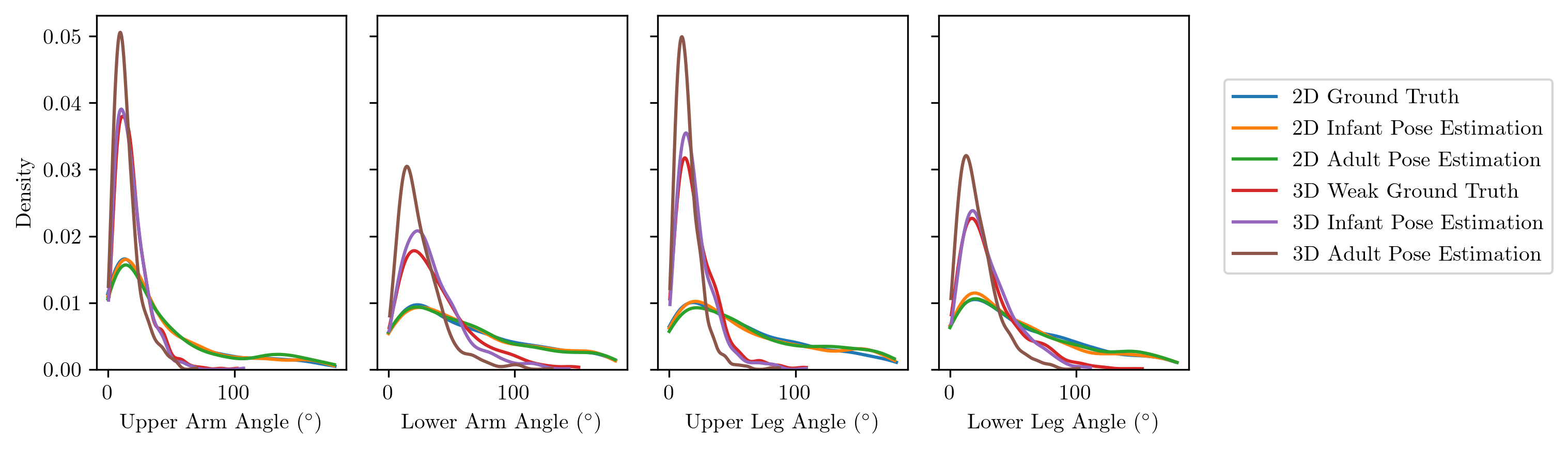}
    \vspace{-.11in}
    \caption{Distribution of raw angle differences, across four pairs of limbs and 700 real SyRIP infant images, as reported by a range of pose-based models. Models based on 3D poses yield far more consistent and seemingly realistic angles, compared with models based on 2D poses.} 
    \label{fig:angle-histograms}
\end{figure*}
}

\newcommand{\figgallerysup}{
\begin{figure*}[b]
   \centering    
   \includegraphics[width=\linewidth]{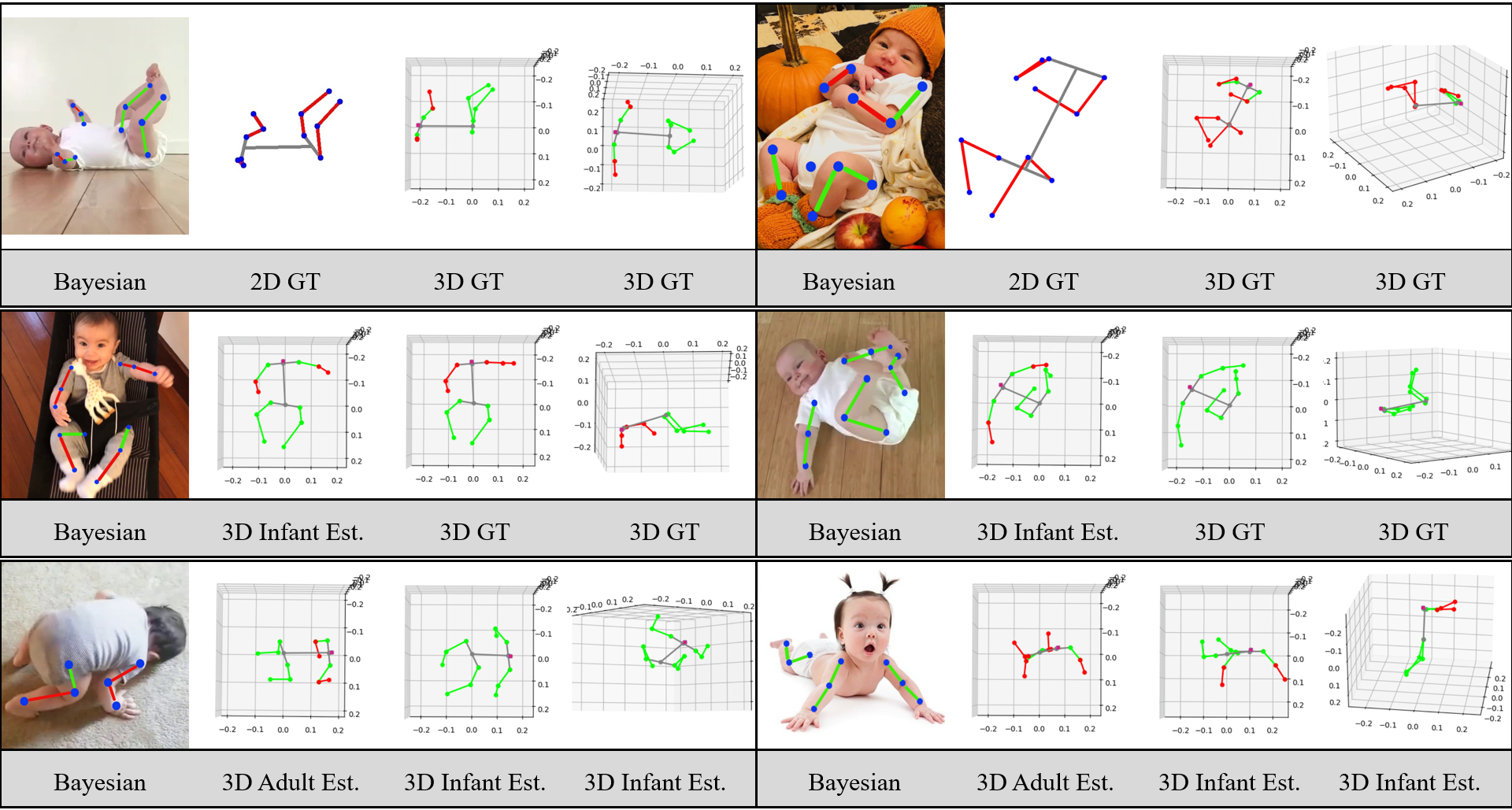}
    \vspace{-.11in}
    \caption{Comparison of the performance of different pose-based symmetry measurement results. In each row we compare symmetry assessments for each pair of limbs from the following models: \textit{1st row:} SyRIP 2D ground truth pose (2D GT) vs. weak 3D ground truth pose (3D GT); \textit{2nd row:} 3D predicted pose by using infant HW-HuP model (3D Infant Est.) vs. weak 3D ground truth pose (3D GT); \textit{3rd row:} 3D predicted pose by using adult SPIN model (3D Adult Est.) vs. 3D predicted pose by using infant HW-HuP model (3D Infant Est.). Bayesian aggregate results are overlaid on the original images, as a kind of weak ground truth. Labeling conventions as in \figref{gallery}. 3D infant pose estimation yields better results than those obtained from the 3D adult pose estimation, but the best results come from the weak 3D ground truth. The Bayesian result of left-side example in the 3rd row is incorrect, possibly, due to the effects of occlusion on human judgement. Our pose-based methods have been trained to be robust to occlusion, and can produce objective evaluations where human assessments falter.
 } 
    \label{fig:gallerysup}
    \vspace{-.2in}
\end{figure*}
}

\newcommand{\tbldatatypes}{
\begin{table*}[t]
\caption{List of data types associated with each infant image considered in this paper. The underlying 700 real infant images are sourced from the synthetic and real infant pose (SyRIP) dataset.} 
\footnotesize
\vspace{-.2in}
\begin{center}
        \begin{tabular}{p{1.9cm}p{2.4cm}p{11.8cm}}
        \toprule
        Type & Data & Source\\ 
        \midrule
        2D body pose & 2D coords. of 17 body joints & Included in \textbf{SyRIP ground truth} (\secref{data}) or inferred from images by \textbf{2D pose estimation} models (\secref{pose-estimators})\\
        \midrule
        3D body pose & 3D coords. of 14 body joints & Our \textbf{weak 3D ground truth} labels for SyRIP (\secref{3d-pose-corr}), or inferred from images by \textbf{3D pose estimation} models (\secref{pose-estimators}) \\
        \midrule
        Pose-based raw angle & Angle in $^\circ$ for 4 limb pairs & \textit{Differences} between the left and right angles for 4 pairs of limbs (upper arms, lower arms, upper legs, lower legs), \textbf{derived from 2D or 3D body pose data} (above), either ground truth or predicted \\
        \midrule
        Pose-based angle class & 3-level angle class for 4 limb pairs & Class from [${<}30^\circ$, $30^\circ$--$59^\circ$, ${\geq}60^\circ$], either \textbf{inferred from pose based raw angles} (above), or canvassed from \textbf{human raters}, (\secref{survey}), or \textbf{voted} or \textbf{Bayesian aggregated} from human raters (\secref{human-bayesian-asssessment}) \\
        \midrule
        Pose-based symmetry & Binary symmetry class for 4 limb pairs & Either inferred from pose based raw angles (above) based on specified threshold angles, or canvassed from \textbf{human raters} (\secref{survey}), or \textbf{voted} or \textbf{Bayesian aggregated} from human raters (\secref{human-bayesian-asssessment}) \\
        \bottomrule
        \end{tabular}
\label{tbl:data-types}
\end{center}
\vspace{-.2in}
\end{table*}
}

\newcommand{\tblestimators}{
\begin{table}[b]
\caption{\textit{Top:} Pose estimation models or ground truth data, from which symmetry assessments are derived. \textit{Middle (see \secref{pose-based-analysis}):} The optimal threshold angle for each pose estimation model for obtaining symmetry assessments with the highest Cohen's $\kappa$ agreement with the Bayesian aggregated symmetry assessments. \textit{Bottom (see \secref{pose-based-analysis}):} Areas under the receiver operating curves (\figref{roc-curve}) for logistic regression of the Bayesian aggregated symmetry assessment using the raw angles obtained from each pose estimation model.}
\vspace{-.2in}
\footnotesize
\begin{center}
\resizebox{\columnwidth}{!}{
        \begin{tabular}{lcccc}
        \toprule
        & & Adult Pose Est. & Infant Pose Est. & Ground Truth \\ 
        \cmidrule{2-5}
        \ldelim\{{2}{*}[Source\hspace{6.9mm}] & 2D & DarkPose & FiDIP & SyRIP    \\  
        & 3D & SPIN & HW-HuP & Corrected HW-HuP \\
        \cmidrule{2-5}
        \ldelim\{{2}{*}[Opt. Angle\hspace{2mm}]& 2D & 43.7 & 42.0 & 39.1 \\  
        & 3D & 17.8 & 27.9 & 27.7 \\
        \cmidrule{2-5}
        \ldelim\{{2}{*}[ROC AUC\hspace{2.5mm}] & 2D & 0.60 & 0.61 & 0.62    \\  
        & 3D & 0.60 & \textbf{0.68} & 0.73 \\
        \bottomrule
        \end{tabular}}
\label{tbl:estimators}
\end{center}
\end{table}
}

\newcommand{\tblbayes}{
\begin{table}[b]
\vspace{-.2in}
\caption{\footnotesize Performance of each human rater assessment of symmetry relative to Bayesian aggregate.} 
\vspace{-.2in}
\begin{center}
\resizebox{\columnwidth}{!}{
        \begin{tabular}{lrrrrrrrrrr}
        \toprule
        Rater \# & 1 & 2 & 3 & 4 & 5 & 6 & 7 & 8 & 9 & 10 \\ 
        \midrule
        Sensitivity           & 0.98   & 0.97   & 0.88   & 0.54   & 0.99   & 0.61   & 0.99   & 0.64   & 0.38   & 0.79    \\ 
        Specificity           & 0.35   & 0.38   & 0.80   & 0.96   & 0.03   & 0.96   & 0.30   & 0.89   & 0.98   & 0.80    \\ 
        \bottomrule
        \end{tabular}
        }
\label{tbl:bayes}
\end{center}
\end{table}
}

\newcommand{\tblpose}{
\begin{table}[ht]
\caption{3D pose estimation performance in mean per joint position error (MPJPE) in mm  on the 100 image SyRIP test set with our weak ground truth 3D labels. We compare our fine-tuned HW-HuP model (HW-HuP-FT) with the base HW-HuP model, and with the adult-trained SPIN model.} 
\footnotesize
\vspace{-0.2in}
\begin{center}
\resizebox{0.7\columnwidth}{!}{
\begin{tabular}{lrrr}
\toprule
Model & SPIN  & HW-HuP & HW-HuP-FT\\
\midrule
MPJPE & 105.8 & 97.2 & \textbf{78.3}              \\ 
\bottomrule
\end{tabular}}
    \label{tbl:pose}
\end{center}
\vspace{-0.2in}
\end{table}
}

\newcommand{\tblfactor}{
\vspace{-.1in}
\begin{table}[h]
\caption{\footnotesize Parameter importance evaluation for logistic regression of Bayesian aggregate symmetry class from human rater and 3D infant pose estimate assessments.} 
\vspace{-.1in}
\resizebox{\columnwidth}{!}{
\begin{tabular}{lrrrrrrr}
\toprule
  && \multicolumn{2}{c}{Human Raters}   && \multicolumn{2}{c}{3D Infant Pose Est.} \\
\cmidrule{3-4} \cmidrule{6-7}
Excluded Feature&& $R_{\text{CS}}^2$ & $\Delta R_{\text{CS}}^2$ && $R_{\text{CS}}^2$      & $\Delta R_{\text{CS}}^2$      \\ 
\midrule
None (Full Model)        && 0.408         &             && 0.455              &                  \\
Limb Part && 0.402         & -0.006      && 0.418              & -0.037           \\
Posture  && 0.408         & 0.000           && 0.448              & -0.007           \\
Angle  && 0.066         & -0.342      && 0.089              & -0.366           \\
Occlusion && 0.408         & 0.000      && 0.455              & 0.000                \\ 
\bottomrule
\end{tabular}
}
\vspace{-.2in}
\label{tbl:factor}
\end{table}
}

\newcommand{\tbllinearregression}{
\vspace{-.1in}
\begin{table*}[p]
\begin{center}
\caption{\footnotesize Proportion of variance $R^2$ values from three linear regression models, all with the 3D weak ground truth raw angle as the dependent variable (and each of four pairs of limbs across 700 infants as the sample space). All three models include the labels for the limb part, the posture, and occlusion as independent variables, together with an angle class label drawn respectively the 2D infant pose estimation, 3D infant pose estimation, or the Bayesian aggregate assessment. The resulting $R^2$ and adjusted $R^2$ scores can be interpreted as gauging the predictive power of each respective model, with our 3D infant pose estimation method holding a clear advantage.}
\resizebox{0.3\columnwidth}{!}{
\begin{tabular}{lrr}
\toprule
Model & $R^2$  & Adjusted $R^2$ \\
\midrule
Bayesian Aggregate                   & 0.365    & 0.360               \\
2D Infant Pose Estimation              & 0.229    & 0.223              \\
3D Infant Pose Estimation              & \textbf{0.496}    & \textbf{0.492}  \\ 
\bottomrule
\end{tabular}}
\label{tbl:linearregression}

\end{center}
\end{table*}
}

\newcommand{\supp}{
\newpage
\onecolumn

\section{Supplementary Materials}

In this section we offer figures and analyses supplementing our discussion from the main paper. \figref{rater-thresholds} and \figref{angle-histograms} provide overviews of the human rater survey data, and the pose-based angle assessments, respectively. \figref{gallerysup} shows further skeleton visualizations of human and machine assessments of symmetry, highlighting comparative differences. Finally, \tblref{linearregression} contains one more regression variance analysis. These results further reinforce our findings regarding the advantages of 3D pose-based symmetry assessment over both human ratings (individual or in aggregate) and 2D pose-based systems.

\figraterthresholds
\figanglehistograms
\figgallerysup
\tbllinearregression
}

\def\wacvPaperID{110} 

\wacvapplicationstrack 

\wacvfinalcopy 

\ifwacvfinal
\usepackage[breaklinks=true,bookmarks=false]{hyperref}
\else
\usepackage[pagebackref=true,breaklinks=true,colorlinks,bookmarks=false]{hyperref}
\fi

\begin{document}

\title{Computer Vision to the Rescue: Infant Postural Symmetry Estimation from Incongruent Annotations}

\author{Xiaofei Huang$^{1}$, Michael Wan$^{1}$, Lingfei Luan$^{1}$, Bethany Tunik$^{2}$, Sarah Ostadabbas$^{1*}$\\
$^{1}$Augmented Cognition Lab (ACLab), Northeastern University, Boston, MA, USA\\
$^{2}$Board-Certified Clinical Specialist in Pediatric Physical Therapy\\
{$^*$Corresponding author's email: \tt\small ostadabbas@ece.neu.edu}
}

\maketitle
\thispagestyle{empty}


\begin{abstract}
Bilateral postural symmetry plays a key role as a potential risk marker for autism spectrum disorder (ASD) and as a symptom of congenital muscular torticollis (CMT) in infants, but current methods of assessing symmetry require laborious clinical expert assessments. In this paper, we develop a computer vision based infant symmetry assessment system, leveraging 3D human pose estimation for infants. Evaluation and calibration of our system against ground truth assessments is complicated by our findings from a survey of human ratings of angle and symmetry, that such ratings exhibit low inter-rater reliability. To rectify this, we develop a Bayesian estimator of the ground truth derived from a probabilistic graphical model of fallible human raters. We show that the 3D infant pose estimation model can achieve 68\% area under the receiver operating characteristic curve performance in predicting the Bayesian aggregate labels, compared to only 61\% from a 2D infant pose estimation model and 60\% from a 3D adult pose estimation model, highlighting the importance of 3D poses and infant domain knowledge in assessing infant body symmetry. Our survey analysis also suggests that human ratings are susceptible to higher levels of bias and inconsistency, and hence our final 3D pose-based symmetry assessment system is calibrated but not directly supervised by Bayesian aggregate human ratings, yielding higher levels of consistency and lower levels of inter-limb assessment bias\footnote{Dataset and model code available at \href{https://github.com/ostadabbas/Infant-Postural-Symmetry}{https://github.com/ostadabbas/Infant-Postural-Symmetry}.}. 
\end{abstract}


\section{Introduction}
\label{sec:intro}


Persistent asymmetrical body behavior in early life provides a prominent prodromal risk marker of neurodevelopmental conditions like autism spectrum disorder (ASD), which affects about 2\% of children \cite{teitelbaum1998movement,Teitelbaum11909,esposito2009exploration,esposito2009symmetry}.
It is also symptomatic of congenital muscular torticollis (CMT), a common musculoskeletal condition with an estimated incidence of 3.9\% to 16\% of infants \cite{kaplan_physical_2018}. Early screening of ASD and CMT is critical for timely intervention and support \cite{sargent2019congenital}, but currently requires laborious professional behavioral assessments, and for ASD, reliable determinations often only come later in childhood. In this paper, we propose a computer vision method for assessing bilateral infant postural symmetry from images, based on 3D human pose estimation, domain adapted to the challenging setting of infant bodies. Our method appears to be less susceptible to inter-limb biases present in human ratings, and as such could be used to great effect in telehealth, where even experts might find it difficult to judge 3D symmetry from on-screen 2D images. Since our system is based on angles extracted from pose estimation, it is both privacy-preserving and highly interpretable, and is adaptable to new definitions of postural symmetry based on updated scientific hypotheses or discoveries, as well as for different conditions.

Our model assesses bilateral postural asymmetry, first by employing state-of-the-art 3D body pose  estimation designed specifically for infant bodies, and second by learning a pose-based assessment calibrated to human ratings of asymmetry. The pipeline is simple but its implementation is highly nontrivial because \textit{reliable ground truth data does not exist for either task}. For pose estimation, there are no infant datasets labeled with 3D ground truth poses, which would require apparatus infeasible for infant subjects. We make some headway by expanding an existing infant body dataset with new 3D pose labels obtained by manual correction of predictions attained from a 3D infant pose estimation model. Nonetheless, as this 3D data is guided only by perception from flat images, these labels can only serve as weak 3D ground truth. As for symmetry assessment, we conduct a pioneering survey of 10 human raters for their assessments of pose symmetry and angle differences in four pairs of limb across 700 infant images, and find low inter-rater reliability, and suggestions of low internal consistency and high bias. In both settings, ground truth data is constrained by the fundamental challenge of deriving three-dimensional information from two-dimensional images, especially in the domain of infant bodies. 

The chief technical thrust in this paper is to ``bootstrap'' from kernels of reliable information in both tasks to obtain globally reliable and bias-free computer vision assessments of body symmetry. Specifically, our strategy is as follows:

\figgallery
\begin{itemize}[leftmargin=*,topsep=0pt,itemsep=-0.33ex,partopsep=0.33ex,parsep=0.33ex]
    \item To remedy the lackluster human rater reliability, we employ a probabilistic graphical model of the human raters as fallible assessors, and compute a Bayesian aggregate of the underlying ground truth, which appears to exhibit a higher level of internal consistency than the human ratings it is derived from. (Full reliability is not assumed.)
    \item We show that for infant images, the body joint angles obtained from the infant 3D pose estimation model can be used alone to predict the Bayesian ground truth assessments on those images with reasonable accuracy, about 68\% area under the receiver operating characteristic (ROC) curve. By comparison, angles obtained from an infant 2D pose estimation model only achieves  61\% area under the ROC curve. Some visualized examples of this discrepancy are shown in \figref{gallery}. The power to predict a response variable obtained completely independently, and despite potential noise in both variables, provides evidence of the accuracy of the 3D pose estimated angles. 
    \item Finally, we learn a simple symmetry classifier for infant images based on the 3D pose estimated angles---now known to be fairly accurate---calibrated by the Bayesian aggregate symmetry rating. This final classifier is \textit{guided} by human intuitions of cutoff thresholds for symmetry assessment, but by design and as we verify quantitatively, is free from the apparent biases stemming from errant factors which affect human judgement. We also demonstrate its superiority over an analogous classifier derived from 2D pose estimates.
\end{itemize}

, yielding our final product of an end-to-end 3D pose-based symmetry assessment which emulates human judgements, and is at the same time demonstrably less susceptible to human biases unearthed by our human survey analysis.

Altogether, we offer an exploration of the challenges of human and machine-learning assessment of human body symmetry, and distill our insights into an adaptable, interpretable, end-to-end algorithm for assessing infant symmetry from still images, with a view towards applications to the early detection and treatment of ASD, CMT, and other common neurodevelopmental conditions. 

\section{Related Work} 
Over the past decade, computer vision has been used in the field of automated medical diagnosis, including to distinguish atypical development through video-based behavior monitoring \cite{thevenot2017survey,de2020computer}. These vision-based systems provide a low-cost and non-invasive approach, developing a more objective way to analyze data, and potentially reducing healthcare expenditures when compared to medical examinations. Among key vision-based biomarkers is persistent body asymmetry in infants, which indicates abnormalities associated with developmental disorders, such as ASD and CMT \cite{esposito2009symmetry,sargent2019congenital}. 

In \cite{sarmiento2020behavioral}, as part of the behavioral phenotyping for ASD, authors examined the arm movement and asymmetry in children. They extracted arm and shoulder angles of the child from recorded videos, using a pre-trained real-time multi-person 2D pose estimation model, OpenPose \cite{cao2017realtime}. A computer vision tool to measure and identify ASD behavioral markers based on components of the autism observation was introduced in \cite{hashemi2012computer}. Authors first applied 2D pose estimation, which is proposed by extending the Object Cloud Model (OCM) segmentation framework \cite{miranda2010cloud} to work with video data, and to produce a 2D stick-man of the toddlers in video segments in which they were walking naturally. Then static and dynamic arm symmetry, as one type of the behavior marker, was detected using the absolute 2D angle difference between corresponding arm parts across time in video segments. Asymmetry was defined if the angle between two corresponding arm parts differs by more than $45^{\circ}$. Both of these papers detected body movement symmetry based on the measured angle differences of arm pairs. 

Meanwhile, in \cite{hocking2022feasibility}, authors developed a virtual reality (VR)-based motor intervention methodology by using motion tracking data to quantify efficiency, synchrony and symmetry of whole-body movement. They proposed another kind of hand bilateral symmetry definition, which is the average and standard deviation of the difference in absolute value of horizontal distance between the hands. For symmetry measurement, the 2D locations of wrists were predicted by the pose estimator integrated in the Microsoft Kinect API and then the symmetry score was calculated according to their proposed symmetry measurement formula. 

A universal shortcoming of all previous computer vision-based approaches to postural symmetry, however, is their reliance on measurements from 2D body poses, even though human body movement and symmetry is fundamentally three-dimensional. Postural symmetry measurement via 3D body poses has yet to be explored.

\section{Concepts and Methods}

\subsection{Pose-Based Symmetry Measurement}
\label{sec:SymMethod}

In this paper, we work with a simple parameterized measurement of symmetry for body limbs based on 2D or 3D body joint locations, inspired by the definition of pose symmetry in \cite{esposito2009symmetry}. First, the infant 2D or 3D pose or skeleton, a collection of human joint locations, is extracted from a flat image by pose estimation algorithms. There are mature computer vision algorithms for this task, but their performance is weaker in the data-scarce infant domain, so we make use of models adapted specifically to infant bodies. For 2D pose extraction, we use the fine-tuned domain-adapted infant pose (FiDIP) model from \cite{huang2021invariant}, which works by fine-tuning from an adult pose model to the infant domain, leveraging a domain adversarial network to learn equitably from both real and synthetic infant data. For 3D infant pose detection, a heuristic weakly supervised human pose (HW-HuP) estimation approach \cite{liu2021heuristic} is applied. HW-HuP learns partial pose priors from public 3D human pose datasets in flexible modalities, such as RGB, depth or infrared signals, and then iteratively estimates the 3D human pose and shape in the target infant domain in an optimization and regression hybrid cycle. These infant 2D and 3D pose estimators output 17 keypoints and 14 keypoints locations, respectively, but we restrict our poses to the 12 body keypoints needed to define the upper and lower arms and legs (shoulders, elbows, wrists, hips, knees, and ankles), where asymmetry is most prominently manifested.

From the 12 keypoints in the body pose, we can obtain measurements of angles and assessments of symmetry geometrically, as follows, and as illustrated in \figref{symmthd}. First, consider the line segment $l_\text{s}$ connecting the two shoulder joints, and then define its mid-perpendicular $p_\text{s}$, the line (in 2D) or plane (in 3D) which intersects $l_\text{s}$ orthogonally at its midpoint. Then reflect the upper right arm across $p_\text{s}$, shift it so that its shoulder joint is aligned with that of the left upper arm, and measure the resulting angle. Similarly, reflect the right forearm across $p_\text{s}$, shift it so that its elbow joint is aligned with that of the left forearm, and measure the angle. This is repeated for the legs: reflect, align, and compare the right versus left upper and lower leg angles, this time across the mid-perpendicular of the segment $l_\text{h}$ connecting the hip joints. If the formed angle of a given limb pair is less than some fixed predefined angle $\theta$, then the the corresponding limb pair is considered to be symmetric, and otherwise it is asymmetric. 
By adopting above proposed approach and varying angle thresholds, we are able to produce raw angle values and pose symmetry labels for each limb pair in infant images based on their 2D and 3D skeletons.

\figsymmthd
\subsection{Human Symmetry Assessment and Bayesian Aggregation}
\label{sec:human-bayesian-asssessment}

Pose asymmetry is often assessed by clinical experts to gauge neurodevelopment, or as a symptom for certain developmental disorders. To guide our algorithmic efforts in emulating clinical evaluations, we surveyed a number of human raters for their assessments of pose angle differences and symmetry in infant images, for the pairs of limbs from our symmetry measurement described in \secref{SymMethod}. The raters were asked to assess angle differences for limb pairs as per our measurement method, and also to make a subjective judgement of symmetry for each limb pair unguided by this method, to reduce redundancy and to capture information about innate symmetry assessments.

\tbldatatypes
We find that in practise, there is large variation and weak agreement amongst assessments from human raters, and this lack of reliability is not alleviated by simple majority voting, in part because such voting is susceptible to noise from outliers. To rectify this, we employ a probabilistic approach to evaluate different annotators and also give an estimate of the actual hidden labels, as proposed in \cite{raykar2010learning}. When multiple annotators provide possibly noisy labels and there is no absolute gold standard, a maximum-a-posteriori (MAP) estimator is proposed to jointly learn the classifier or regressor, the raters accuracy, and the actual true label. The performance of each rater is measured by calculating sensitivity and specificity with respect to the unknown gold standard, and then a higher weight is assigned to them. We apply an expectation maximization (EM) algorithm to measure the performance of raters based on the given standard, and then optimize the standard based on the new rater performance. The gold standard is initialized by the majority voting result. 

Considering that we want to trust some particular raters more than others, a prior knowledge is imposed to potentially capture the skill of different raters. Beta priors, randomly initialized, are given as conditional information when calculating the probabilities of sensitivity, specificity, and prevalence for our Bayesian aggregation approach. Specifically, following the data gathered in our survey, there are two different types of rating labels: (1) binary class labels, as symmetric or asymmetric, and (2) angle class labels [${<}30^\circ$, $30^\circ$--$59^\circ$, ${\geq}60^\circ$], which intrinsically ordered.
For the first binary symmetry labels, we infer the human true label directly following the EM optimization procedures mentioned above. In terms of the ordinal angle labels, as described in \cite{raykar2010learning}, we first transform it into two new binary class labels, and then applying the same estimation procedure for each binary data to get probability of new label. After that, the probability of the actual class values can be calculated when combine these two transformed binary data. For each instance, we assign the class with the maximum probability.

\section{Annotation from Humans and Machines}
\label{sec:data}

In order to evaluate the performances of human rating and pose-based symmetry measurement, we apply them to a real infant image set of the publicly released synthetic and real infant pose (SyRIP) data \cite{huang2021invariant}, which contains 700 real images with assigned posture labels (supine, prone, sitting, and standing) and annotated 2D keypoint locations. See \tblref{data-types} for an overview of the data types and sources discussed in this paper.

\tblestimators
\figrateragreement
\subsection{Human Symmetry Survey}
\label{sec:survey}

In order to reveal and simulate the mechanism of human rating for postural symmetry, we conducted an online experiment study to collect the pose symmetry judgement responses of SyRIP real images from 10 raters through Qualtrics platform. The 700 images were divided into 28 sections, each of which had 25 questions. The questions in each block were randomly assigned to each participant. There were two sessions of mandatory resting time (5 minutes) assigned after the 10th and 20th sections. Each image was accompanied by eight questions: four of them regarding the symmetry of the four limb pairs (upper arm, lower arm, upper leg, and lower leg) and the rest about the predicted angle class between each of the four pairs of limbs. There were five demographic questions at the end of the survey about their major, gender, age, education level, and experience in computer vision or drawing (23 was the mean age; there were 5 male and 5 female participants). A basic snapshot of the survey responses, which plots the mean rater assessment of symmetry at each assessed angle class, can be found in \figref{rater-thresholds} in the Supplementary Material.

\subsection{Infant 2D and 3D Pose Estimation}
\label{sec:pose-estimators}
We tested the performance from a number of pose estimation models, listed in \tblref{estimators}. We applied DarkPose model \cite{zhang2020distribution}, which is trained on large-scale public human pose datasets, adapted to infant poses using FiDIP model to predict 2D keypoints for 2D pose-based symmetry measurement. The well-performed human 3D pose estimation model, SPIN \cite{kolotouros2019learning}, and the infant-adapted 3D pose estimator, HW-HuP \cite{liu2021heuristic}, are used to infer 3D keypoints for our proposes 3D pose-based symmetry measurement. 2D and 3D pose ground truth come from SyRIP dataset and corrected HW-HuP predictions, respectively.

\subsection{Infant 3D Pose Correction}
\label{sec:3d-pose-corr}
The performance of our proposed pose-based symmetry measurement depends largely on the accuracy of the 2D or 3D pose estimation. The SyRIP dataset, however, only contains ground truth keypoint locations in 2D coordinates, not 3D. To fill this gap, we modified the interactive annotation tool introduced in \cite{Li_2020_CVPR} to correct poses predicted by the infant 3D pose estimator, HW-HuP. Since this pose estimator also estimates camera parameters, we overlay its 3D pose keypoint predictions onto the 2D plane over the original infant image, to ensure 2D pose alignment. We interactively modify the global pose orientation and the local bone vector orientation of the 3D skeleton by keyboard inputs to make both the 3D skeleton and the real-time updated projected 2D keypoints locations as correct as possible. In this way, we obtained the \textit{weak ground truth of 3D pose} because of the error of inevitable from human vision and camera parameter estimation. The distributions of predicted angle differences obtained from various 2D and 3D pose estimation models or ground truth are exhibited in \figref{angle-histograms} in the Supplementary Material.

\section{Analysis: Computer Vision to the Rescue}

We start our analysis by examining shortcomings of human ratings of symmetry, and illustrate how our Bayesian aggregation process ameliorates some of these issues. We then demonstrate the ability of the 3D infant pose estimation models to predict the Bayesian aggregate assessments of angle and symmetry to a higher degree than adult or 2D pose-based models, increasing our confidence in both the Bayesian aggregates and the 3D pose estimations. Finally, we produce our end-to-end symmetry assessments by calibrating the 3D pose-based symmetry assessments with the Bayesian aggregate data. We demonstrate performance gains afforded by the 3D infant pose-based system over 2D or adult pose-based alternatives, and also illustrate the advantages offered by our algorithmic pose-based assessments over the human and even Bayesian aggregate assessments, both quantitatively and qualitatively. We round out our analysis with two codas on 3D pose estimation and factors affecting symmetry assessment. 

\subsection{Amending Incongruent Human Annotations}
\label{sec:rater-reliability}

The average Cohen's $\kappa$ agreement\footnote{Cohen's $\kappa$ measures rater agreement on a scale from -1 to 1, while (unlike correlation) accounting for agreements due to random chance. It is often interpreted as a measure of \textit{inter-rater reliability}.} between each human rater and their other nine fellow human raters, on their assessments of angle class and symmetry across four pairs of limbs and 700 real images in the SyRIP infant dataset, can be found in  \figref{rater-agreement}. It attests to generally ``fair'' average agreement for angle class assessments and ``slight'' to ``fair'' average agreement for symmetry assessments. In the same vein, the Krippendorff's $\alpha$ collective agreement amongst the entire group of human raters is 0.30 for angle class and 0.18 for symmetry, attesting respectively to ``fair'' and ``poor'' collective agreement. In addition to lower inter-rater agreement, human assessments are also afflicted with high inter-limb assessments agreement for angle class and especially symmetry, as seen in \figref{part-correlations}. The high arm-to-leg agreement in symmetry assessments in particular likely indicate unwarranted bias, given that corresponding arm-to-leg agreement for angle class are negligible. Finally, \figref{internal-consistency} shows that human ratings of angle class exhibit a low level of correspondence with human ratings of symmetry, suggesting a low level of ``internal consistency'' among individual human ratings.

\figpartcorrelations
\figinternalconsistency

Underlying many of these issues is the high variance in assessments between raters (illustrated starkly in the precis of individual rater responses in \figref{rater-thresholds} in the Supplementary Material), and the high variance of the resulting agreement and consistency metrics. These issues prompt us to explore methods of aggregating the human ratings into a more cohesive whole, including a simple voting method and the probabilistic Bayesian aggregation method described in \secref{human-bayesian-asssessment}. The results in \figref{rater-agreement}, \figref{part-correlations}, and \figref{internal-consistency} corresponding to these aggregation methods show that the Bayesian aggregate in particular enjoys lower inter-limb agreement and higher angle-asymmetry correspondence than the average human rater---suggesting respectively, lower levels of bias and higher internal consistency---all while maintaining a high level of agreement with the average human rater. Thus, we adopt the Bayesian aggregate assessment as a weak ground truth representation of human assessment of symmetry, with potentially undesirable characteristics excised. \tblref{bayes} reports performance metrics of individual human assessments of symmetry, relative to the Bayesian aggregate results as ground truth, again demonstrating the wide variance in human reliability.

\tblbayes
\subsection{Pose-Based Symmetry Assessment}
\label{sec:pose-based-analysis}

\figroccurve

As promised, we now analyze the extent to which the pose-based systems can track the Bayesian aggregate symmetry assessments, and then calibrate and evaluate our final automated symmetry assessment system.
We first consider the raw angle data obtained from pose estimation, which, as described in \secref{SymMethod}, consists of set of four angle differences, in degrees, for the four key pairs of limbs under consideration (upper arms, lower arms, upper legs, and lower legs). This will soon be converted to the discrete signal of the angle category and symmetry assessment, but for now we retain the maximum amount of information and gauge agreement with the Bayesian aggregate rater assessments of angle class and of symmetry for each of the four joints and each infant image (2800 data points in all) by logistically regressing for them using the raw angles. \figref{roc-curve} shows the receiver operating characteristic (ROC) curves resulting from this regression, performed with a 3:1 train-test split. For the regression of angle class, we compress the three true classes into a binary variable indicating whether the angle is over $30^{\circ}$ for ease of interpretation. The areas under the curve (AUC) for the ROC curves in the symmetry regression are provided in \tblref{estimators}. 

\figoptimalangle
\figtypical

These metrics confirm that the raw angles from the weak 3D ground truth can model the Bayesian aggregate assessment of both angle and symmetry to a high degree of fidelity. Neither set of data can be taken as fully reliable ground truth, but since they are derived from different human annotators performing fairly different tasks, the high level of agreement exhibited here increases our confidence in the accuracy of both. Among pose estimation models, the 3D infant-specific models enable the next best predictions of human symmetry assessments, while the poses from the remaining models---either 3D pose models for the general most adult human, or 2D pose models for infants or adults---offer weaker ability to predict the human assessments.

We now turn to the task of calibrating an end-to-end pose-based system for the evaluation of symmetry, for use in practical applications or further research\footnote{For some applications, the output of the four raw angles will suffice as a multi-dimensional, continuous measure of overall body symmetry, affording the end-user the flexibility to redefine the overall notion of symmetry for different tasks or in response to advances in scientific understanding.}. In concrete terms, wish to select threshold angles which will allow us to convert our raw joint angles to binary assessments of symmetry per joint, in a way that maximizes concordance with the Bayesian aggregate. We choose to guide this concordance with the same Cohen's $\kappa$ agreement score employed earlier. \figref{optimal-angle} shows the Cohen's $\kappa$ agreement of symmetry assessments derived from all six of the pose-based estimators at various decision angle thresholds, compared with both the voted and Bayesian raters; it also shows the mean Cohen's $\kappa$ agreement with each of the ten human raters individually. Incidentally, these results not only confirm again the supremacy of the 3D ground truth and infant pose prediction methods for tracking human assessments of symmetry, but on the flip side, also demonstrate the superiority of the Bayesian aggregation of human symmetry assessments over the voted or the average human assessment for tracking the 3D weak ground truth assessment, at most reasonable angle thresholds. From these Cohen's $\kappa$ curves, we extract the threshold angles maximizing agreement for each pose-based model, reported in \tblref{estimators}. These thresholds then define the corresponding symmetry assessment for each model (or ground truth data).

Metrics quantifying these final assessment models have already been reported throughout the paper, but we offer our interpretations here. \figref{rater-agreement}, confirms, as expected, that the 3D infant pose estimation assessment offers the highest average agreement with human rater assessment, compared to the 3D adult pose estimation or the 2D infant pose estimation models; our model also comes close to the level achieved by the 3D weak ground truth assessment. \figref{part-correlations} shows that assessments based on predicted or weak ground truth 3D poses are relatively free from inter-limb agreement, compared to individual or aggregate human assessments. In the absence of fully reliable ground truth assessments, this circumstantially suggests that human assessments are susceptible to bias from nearby parts, while our automated approach is not. Finally, \figref{internal-consistency} shows that most of the pose estimation based models enjoy high internal consistency in their assessment of angle class versus symmetry, as to be expected from mechanistic models.

\subsection{Qualitative Evaluation}
\label{sec:eval}

We illustrate the performance of the pose-based models in \figref{gallery} and \figref{typical}. These figures visualize the Bayesian aggregate assessments of symmetry on top of the original image, as well as the assessments derived from the 2D and 3D infant pose models (FiDIP and HW-HuP, respectively) on top of their respective predicted pose skeletons, with green indicating symmetric judgements and red indicating asymmetric judgements. 

In \figref{gallery}, we see examples of infant poses where the 2D pose-based assessment is mistaken, but the 3D pose-based assessment is able to make the correct call, according to the Bayesian aggregate label. We have highlighted multiple views of the 3D skeleton to highlight the advantage that the 3D pose-based assessment has, and suggest that the mistakes made by the 2D pose-based assessment are in a way understandable, given that it is limited to a single perspective, so to speak. \figref{typical} shows special case where the 2D perspective is actually not too limiting, since the infant is lying flat on its back, so its limbs largely confined to a plane parallel to the image plane. Indeed, in this case, the 2D and 3D pose-based symmetry assessments agree, but they differ from the Bayesian assessment, which may reflect human bias or simply a stricter subjective threshold for symmetry on the part of the human assessments. More comparisons between 2D or 3D pose-based assessments are exhibited in \figref{gallerysup} in the Supplementary Material.

\subsection{Coda 1: Improving 3D Pose Estimation}

The main factor limiting performance of state-of-the-art infant 3D pose estimators such as HW-HuP is the scarcity of ``true'' ground truth 3D pose data. We briefly report on the effect of fine-tuning HW-HuP with the weak 3D ground truth labels for SyRIP images generated for this paper. We split the 700 SyRIP images into a 100 image test set, coinciding with SyRIP's Test100 set, and a 600 image train set, and fine-tune the infant HW-HuP model on the 600 image train set with weak 3D labels, for 200 epochs with learning rate of $5e-05$. The resulting performance of the fine-tuned infant HW-HuP model under the mean per joint position error (MPJPE), as reported in \tblref{pose}, is significantly improved over the base infant HW-HuP model, and over the adult pose SPIN model.

\tblpose


\subsection{Coda 2: Factors Affecting Symmetry Assessment}

We conclude our study with a supplementary importance analysis of factors affecting the assessments of symmetry considered in our work, via logistic regression. We take the Bayesian aggregate symmetry assessment as the response variable, and the following as covariate factors: limb part under consideration (upper arm, lower arm, upper leg, or lower leg), the infant posture (included in SyRIP), an occlusion label for each limb (which we annotate for this purpose), and finally, an angle variable. We consider two separate sources for the angle variable, one consisting of the angle class assessments from all of the human raters, and one obtained from the 3D infant pose estimation. We find that both of regressed models are statistically significant. 

According to the logistic regression result for Bayesian aggregation, all four predictors account for between $40.8\%$ ($R_{\text{CS}}^2$) and $54.6\%$ ($R_{\text{N}}^2$) of the variance in the dependent variable and correctly classify $83.2\%$ of cases. From the logistic regression result for four predictors, we conclude that only limb part and estimated angle between corresponding limb part significantly contributed to the asymmetry assessment model. While for 3D prediction model, the logistic regression result indicated that all factors expect occlusion significantly contribute to the model. All four factors explain for between $45.5\%$ ($R_{\text{CS}}^2$) and $60.5\%$ ($R_{\text{N}}^2$) of the variance in the dependent variable and correctly classify $90.8\%$ of cases.

In order to assess predictor importance, we use the decrease of $R_{\text{CS}}^2$ approach to calculate the $\Delta R_{\text{CS}}^2$ when removing one of the predictor. A larger decrease indicates more contribution of the removed predictor to explain the model. The results of decreasing $R_{\text{CS}}^2$ are reported in \tblref{factor}. When the angle estimation was taken out of the model, the $R_{\text{CS}}^2$ value declined by -0.342 for the Bayesian model and by -0.366 for the 3D prediction model correspondingly. Thus, angle estimate is the most important predictor of all the variables that have been found, and it is used in both the human rating model and the 3D prediction model.


\tblfactor


\section{Conclusion}
\label{sec:conclusion}
We have presented a computer vision based method for assessment of postural symmetry in infants from their images, with the goal of enabling early detection and timely treatment of issues related to infant motor and neural development. We found human ratings of symmetry to be unreliable, and rectified them with a Bayesian-based probabilistic aggregate rating. We demonstrated that automatic assessments based on pose estimation avoid some of the pitfalls of human assessments, while retaining the ability to predict the Bayesian aggregate ratings to a strong degree, with 3D infant pose models performing stronger than 2D models or adult pose models.

{\small
\bibliographystyle{ieee_fullname}
\bibliography{ref}
\newpage
\supp
}

\end{document}